\documentclass[letterpaper]{article} 
\usepackage{aaai2026}
\usepackage{times}  
\usepackage{helvet}  
\usepackage{courier}  
\usepackage[hyphens]{url}  
\usepackage{graphicx} 
\usepackage{natbib}  
\usepackage{caption} 
\usepackage{algorithm}
\usepackage{algorithmic}
\usepackage{booktabs}
\usepackage{amsmath}
\usepackage{amssymb}
\usepackage{comment}

\usepackage{multirow}
\usepackage{enumitem}
\usepackage{tabularx}
\usepackage{framed}
\usepackage{amsthm}
\usepackage{amsmath}
\usepackage{amssymb}
\usepackage{xspace}
\usepackage{xcolor}
\usepackage{graphicx}
\usepackage[table]{xcolor}
\usepackage{colortbl}

\newtheorem{thm}{Theorem}

\usepackage{newfloat}
\usepackage{listings}
\DeclareCaptionStyle{ruled}{labelfont=normalfont,labelsep=colon,strut=off} 
\floatstyle{ruled}
\newfloat{listing}{tb}{lst}{}
\floatname{listing}{Listing}
\author{
    Miao Shang, 
    Xiaopeng Hong\thanks{Corresponding author.}
}
\affiliations{

    The Faculty of Computing, Harbin Institute of Technology
    
    miaos0522@gmail.com, 
    hongxiaopeng@ieee.org
    
}

\usepackage{bibentry}

\usepackage{cleveref}
\usepackage{xcolor}
\usepackage{colortbl}
\definecolor{de_blue}{rgb}{0.19, 0.55, 0.91}

\title{2D Gaussians Spatial Transport for Point-supervised Density Regression}
\begin{document}

\maketitle

\begin{abstract}
This paper introduces Gaussian Spatial Transport (GST), a novel framework that leverages Gaussian splatting to facilitate transport from the probability measure in the image coordinate space to the annotation map. We propose a Gaussian splatting-based method to estimate pixel-annotation correspondence, which is then used to compute a transport plan derived from Bayesian probability. To integrate the resulting transport plan into standard network optimization in typical computer vision tasks, we derive a loss function that measures discrepancy after transport. Extensive experiments on representative computer vision tasks, including crowd counting and landmark detection, validate the effectiveness of our approach. Compared to conventional optimal transport schemes, GST eliminates iterative transport plan computation during training, significantly improving efficiency. Code is available at \url{https://github.com/infinite0522/GST}.
\end{abstract}
\section{Introduction}
\label{sec:1}

Optimal transport (OT) has gained great attention in computer vision due to its powerful ability to model and solve various problems involving distances and distributions~\cite{peyre2019computational,villani2021topicsot, wang2020dm,lin2023optimalsemi,zhang2024mahalanobis, ge2021ota}. By treating the prediction and the ground truth as probability distributions, OT determines their optimal match with the transport cost measured by, for example, the Wasserstein distance~\cite{kantorovich1960mathematical}. 

However, despite their effectiveness, the application of OT in deep learning is plagued by high computational costs. This issue stems from their inherent bi-level optimization nature: at each training iteration, an inner loop must solve a costly OT problem (e.g., via the Sinkhorn algorithm~\cite{cuturi2013sinkhorn}) to compute the loss, before an outer loop can update the network's parameters. This tight coupling makes the entire optimization process extremely time-consuming.

To address the computational bottleneck, the primary idea of this paper is to decouple the transport plan generation from the network optimization. The goal is to pre-compute a fixed transport pattern based on the input image and its annotations, and then use it throughout the training. However, this seemingly simple idea poses significant challenges.

First, standard \emph{Optimal Transport} lacks a mechanism to derive a transport plan from the input image prior to network optimization. OT treats the estimated density maps as probability measures in the image coordinate space and solves for the optimal plan to the annotation space. As a result, it requires the network to first generate the density map, followed by the computation of the transport plan and cost.
Second, in practical applications such as crowd counting, the images often contain thousands of targets. To minimize annotation costs, only object locations are typically provided without detailed contours~ \cite{cao2018scale,liu2019context, lin2025semi}. However, object contours are crucial for determining the correspondence between pixels and annotation points, which directly impacts the transport plan. Without contour information, accurately establishing pixel-to-object correspondence becomes challenging.

In response to these challenges, we propose Gaussian Spatial Transport~(GST), a novel framework that efficiently transports probability measures from image coordinates to annotation space. 
At its core, GST establishes an interpretable probabilistic correspondence between pixels and annotations by leveraging 2D Gaussian Splatting. This correspondence is then distilled into a fixed transport kernel, a matrix that encodes the static mapping from each pixel to all target annotations, which can be pre-computed before training. During optimization, the model's predicted density map is pushed forward to the space of ground truth annotations via a single matrix multiplication with this pre-computed kernel.
Whereas OT-based methods require iteratively solving a costly optimization problem to determine the transport plan and compute the loss, our loss is simply the discrepancy between the transported density and the ground truth. Extensive experiments demonstrate that GST achieves a strong balance of high accuracy and computational efficiency.

The main contributions of this paper are threefold:

\begin{figure*}[ht]
\centering
\includegraphics[width=0.9\linewidth]{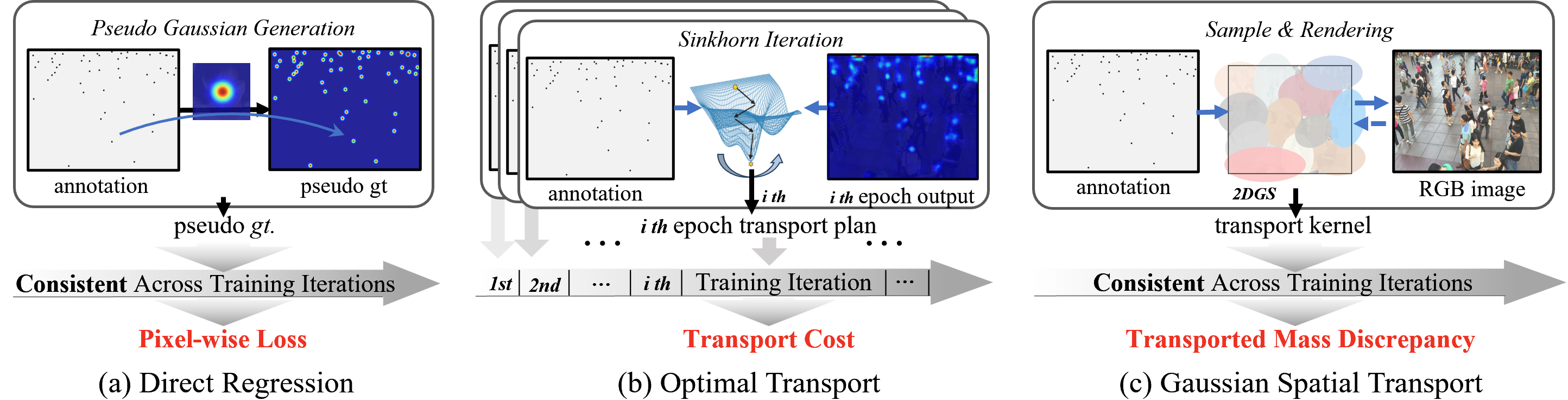}
\caption{Comparison of loss computing of Direct Regression, OT, and the proposed GST.}
\label{fig: intro}
\end{figure*}

\begin{itemize}[noitemsep,topsep=0pt,parsep=2pt,partopsep=0pt,leftmargin=1em]

    \item 
    We propose a novel spatial transport framework that decouples plan generation from network optimization. It relies on a pre-computed transport kernel, which transforms the costly iterative optimization of OT into a single, efficient matrix multiplication during training.
    \item We bridge Gaussian splatting and spatial transport by deriving a Bayesian framework to compute the transport kernel using results obtained from 2D Gaussian splatting.
    \item We successfully apply the proposed GST framework to representative computer vision tasks, including crowd counting and landmark detection.
\end{itemize}

\section{Related work}\label{sec:2}
\subsection{Spatial Transport}

Conventional point-supervised density regression methods~\cite{sindagi2017generating,li2018csrnet, sun2019hrnet} rely on handcrafted Gaussian kernels to generate pseudo-ground-truth density maps for direct pixel-to-pixel regression, creating a critical dependency on the quality of these synthetic approximations, as in Fig.~\ref{fig: intro}(a). Although subsequent improvements~\cite{shanghai,ucfqnrf,wan2020kernel} introduce adaptive kernels to enhance quality, they still suffer from inherent localization ambiguity in high-density scenarios.

Recent works~\cite{wang2020dm,qu2022OTHPE} mitigate these limitations by reframing the task as distribution matching via transport theory. These methods, based on Optimal Transport, treat the predicted density map and the ground truth as two measures, and define the training loss through the minimal cost to transport one to the other. For instance, DMCount~\cite{wang2020dm} first introduced this using balanced OT, while subsequent works explored unbalanced OT to relax the strict mass conservation constraint~\cite{ma2021learning, lin2021direct, wan2021generalized}.
However, it introduces a significant computational bottleneck due to its inherent bi-level optimization nature: at each training step, an inner-loop optimization (e.g., using the Sinkhorn algorithm) must be solved to find the optimal transport plan, before the outer-loop can update the network's parameters. This iterative re-computation of the transport plan makes the entire process time-consuming. Through Bayesian Loss~\cite{ma2019bayesian} avoids this cost by using a fixed, handcrafted transport plan, the reliance on a task-specific prior limits its adaptability.

 Our work addresses the limitations of both. We propose a novel transport scheme, which decouples the transport plan from the network optimization. Instead of using a handcrafted prior, we pre-compute a transport kernel by analyzing the input image's content via Gaussian Splatting. During training, the transport is reduced to a single, efficient matrix multiplication, enabling high performance with significant computational gains. Fig.~\ref{fig: intro} illustrates the key differences between our method and other major paradigms like direct regression and OT.  

\subsection{Gaussian Splatting}

Gaussian splatting (GS) is widely used for efficient appearance modeling and rendering. 
It approximates complex textures and colors of 3D scenes with overlapping Gaussian balls.
Benefiting from the explicit differentiable rendering and GPU-parallelized, tiles-based rasterization pipeline, GS has become a new paradigm for fast differentiable rendering, making it a popular choice for real-time 3D scene reconstruction and editing~\cite{3dgs, huang2024dreamphysics}.

Recently, GS's application has expanded to 2D tasks like image reconstruction~\cite{zhu2025large,dong2025gaussiantoken, ye2024gsplatopensourcelibrarygaussian}, compression~\cite{gaussianimage,wang2025gsvc}, and super-resolution~\cite{hu2024gaussiansr, chen2025generalized}. In the 2D image domain, Gaussian units can serve as novel representation units, replacing traditional pixels. For instance, GaussianImage~\cite{gaussianimage} innovatively replaces pixels with Gaussian units for image representation, further enhancing efficiency by simplifying the parameterization of Gaussian representations and the computationally intensive $\alpha$-blending process in 2D scenes.

Existing methods primarily utilize 2D GS for image reconstruction and restoration. Instead, our work pioneers a novel application of Gaussian Splatting via its connection to spatial transport. We rely on its implicit geometric encoding ability to establish the pixel-to-target correspondence and formulate a transport plan that maps the measures from coordinate spaces to annotation spaces, which is clearly beyond the scope of previous GS studies.

\section{Method}
\label{sec:method}

In this section, we elaborate on our Gaussian Spatial Transport method. We begin with the problem definition and preliminaries. As the design of the loss function is paramount to address the computational bottleneck of OT-based approaches, we provide a general analysis of transport-based losses, which is distinct to the conventional OT formulation with our proposed Bayesian Transport framework. On this basis, the subsequent sections will then detail the full GST pipeline: the generation of its transport kernel via Gaussian Splatting and the training procedure.

\noindent\textbf{Problem Definition.} Let $\mathcal{X}\!\!=\!\!\{x_i\}_{i=1}^{I}$ be pixel coordinates of an RGB image ${\boldsymbol{\zeta}}_{\text{img}}$ and $\mathcal{Y}\!\!=\!\!\{y_n\}_{n=1}^{N}$ be the discrete ground truth point locations. The density map ${\boldsymbol{\zeta}}_d\! \in\! \mathbb{R}_+^{I}$ assigns non-negative density values on each pixel in $\mathcal{X}$. The annotation map ${\boldsymbol{\zeta}}_g \!\in\!\mathbb{R}_+^{N}$ is defined on $\mathcal{Y}$ with $\zeta_g(y_n)\!\!=\!\!1$ for all $n$, representing the once presence of each object. A regression network $f$ with trainable parameter $\theta$ is adopted to approximate the mapping from input to ideal density map ${\boldsymbol{\zeta}}_d$. The loss function for network training is denoted by $L(\cdot)$.

\noindent\textbf{Preliminaries on Transport.}
To establish a quantitative relationship between the density map and annotation points, we first formalize the concept of transport.

By normalizing the two maps into probability functions ${\boldsymbol{P}}_X=\frac{{\boldsymbol{\zeta}}_d}{||{\boldsymbol{\zeta}}_d||_1}$ and $ {\boldsymbol{P}}_Y=\frac{\boldsymbol{\zeta}_g}{||\boldsymbol{\zeta}_g||_1}$, we define the their probability measures $\boldsymbol{\mu}=\sum_{i=1}^{I}P_{X}(x_i)\delta_{x_i}$ on $\mathcal{X}$ and $\boldsymbol{\nu}=\sum_{n=1}^{N}P_{Y}(y_n)\delta_{y_n}$ on $\mathcal{Y}$, where $\delta_{x_i}$, $\delta_{y_n}$ are Dirac measures.

A \emph{spatial transport} between $\boldsymbol{\mu}$ and $\boldsymbol{\nu}$ is a joint probability measure $\boldsymbol{P}=\sum_{i,n}P(x_i,y_n)\delta_{(x_i,y_n)}$ on $\mathcal{X} \times \mathcal{Y}$ that satisfies the marginal constraints. The admissible set is:
\begin{equation}
    \!\mathcal{U}(\boldsymbol{P\!}_X,\boldsymbol{P\!}_Y) \! \triangleq \! \{\boldsymbol{P} \! \in \! \mathbb{R}^{I \times N}_+ \! : \! \boldsymbol{P}\mathbf{1}_N \! = \! \boldsymbol{P\!}_X, \! \ \boldsymbol{P}'\mathbf{1}_I \! = \! \boldsymbol{P\!}_Y \!\}. \!
    \label{trans. def.}
\end{equation}

\subsection{General Form of Transport Plan-based Loss}
Conceptually, any transport-based optimization loss function combines two key aspects: \emph{transport discrepancy}, which measures the difference between the resulting distribution of transported mass and the desired target distribution, and \emph{transport cost}, referring to the expense required to move mass from source to target. Given the density map $\tilde{{\boldsymbol{\zeta}}}_d^t=f({\boldsymbol{\zeta}}_{\text{img}};\theta_t)$ estimated by $f$ at training iteration $t$ and transport plan $\boldsymbol{P}$, a generic loss function is formulated as:
\begin{equation}
    L(\tilde{{\boldsymbol{\zeta}}}_d^t,\boldsymbol{\zeta}_{g}) = \lambda_1 D\big(\tau(\tilde{{\boldsymbol{\zeta}}}_d^t), {\boldsymbol{\zeta}}_{g}\big) + \lambda_2 W(\boldsymbol{P}), 
    \label{eq:generic_loss}
\end{equation}
where $\tau: \mathcal{X} \rightarrow \mathcal{Y}$ signifies the push-forward movement of the source mass transported from $\mathcal{X}$ to $\mathcal{Y}$.
The loss is a weighted sum of two terms: the transport discrepancy, measured by $D(\cdot)$, and the transport cost, given by $W(\cdot)$, which are balanced by parameters $\lambda_1, \lambda_2 \geq 0$.

Existing Optimal Transport-based methods~\cite{wang2020dm, qu2022OTHPE} define their loss by finding the minimal-cost transport from model prediction to ground truth in each training iteration. We, instead, derive a pre-computable transport scheme, termed \emph{Bayesian Transport}, realizing our loss under this fixed-pattern transport.

\subsubsection{Optimal Transport-based Loss} OT-based methods first solve for the minimal cost transport plan ${\boldsymbol{P}}_*^t$ between the normalized predicted density $\tilde{\boldsymbol{P}}_{X}^t=\frac{\tilde{{\boldsymbol{\zeta}}}_d^t}{||\tilde{{\boldsymbol{\zeta}}}_d^t||_1}$ and the ground truth ${\boldsymbol{P}}_Y$, and define the cost term in Eq.~\ref{eq:generic_loss} under this ${\boldsymbol{P}}_*^t$ :
\begin{equation}
\label{eq:ot_objective}
\begin{aligned}
W({\boldsymbol{P}}_*^t) = \min_{\boldsymbol{P}^t} \langle \mathbf{C}, \boldsymbol{P}^t \rangle, \ \ s.t. \ \boldsymbol{P}^t \in \mathcal{U}(\tilde{\boldsymbol{P}}_{X}^t,{\boldsymbol{P}}_Y),\\
\end{aligned}
\end{equation}
where $\mathbf{C} \in \mathbb{R}_+^{I \times N}$ is the cost matrix. 

For the transport discrepancy term, it's important to note two aspects. since $\boldsymbol{P}_*^t$ is solved for normalized probability distributions, $D\big(\tau(\tilde{\boldsymbol{P}}_{X}^t), {\boldsymbol{P}}_Y\big) \! = \! 0$ due to marginal constraints. However, a mass discrepancy still remains when transporting unnormalized distributions (proportionally by $\boldsymbol{P}_*^t$), defined as $D\big(\tau(\tilde{{\boldsymbol{\zeta}}}_d^t), {\boldsymbol{\zeta}}_{g}\big)\!=\! \big| \|\tilde{{\boldsymbol{\zeta}}}_d^t\|_1 \!-\! \|{\boldsymbol{\zeta}}_{g}\|_1 \big|$ in DMCount~\cite{wang2020dm}, which can be proven equivalent to the $L_1$ norm of the difference between the pushed-forward unnormalized distribution and the ground truth ${\boldsymbol{\zeta}}_{g}$\footnote{For more discussions about the definition of $D(\cdot)$ in other methods, please refer to the \emph{Suppl. A}.}.

This approach leads to a \textbf{bi-level optimization}: the inner loop solves the OT problem in Eq.~\ref{eq:ot_objective} through iterative computation within each training step, while the outer loop updates the task-specific model parameters $\theta$ across iterations. This tight coupling requires recomputing ${\boldsymbol{P}}_*^t$ at every step of training, leading to significant computational overhead.

\subsubsection{Bayesian Transport-based Loss} To address OT's computational overhead, we propose a Bayesian Transport-based Loss. This method derives a pre-computable transport plan from probabilistic principles, decoupling its calculation from the regression model's iterative optimization. Its foundation lies in the following theorem.

\begin{thm}
For probability distributions ${\boldsymbol{P}}_X$ on $\mathcal{X}$ and ${\boldsymbol{P}}_Y$ on $\mathcal{Y}$, there exists a transport plan $\hat{\boldsymbol{P}} \in \mathcal{U}({\boldsymbol{P}}_X, {\boldsymbol{P}}_Y)$ that can be expressed as $\hat{\boldsymbol{P}} = \text{diag}({\boldsymbol{P}}_X) \cdot \boldsymbol{\mathcal{K}}$. Here, $\boldsymbol{\mathcal{K}}$, termed transport kernel, has elements defined as \footnote{The kernel is derived following Bayes' theorem and the law of total probability. Proof can be referred to \emph{Suppl. B}.}: 
\begin{equation}
\boldsymbol{\mathcal{K}}_{i,n}=\frac{P(x_i|y_n) P_Y(y_n)}{\sum_{n=1}^N P(x_i|y_n) P_Y(y_n)}.
\end{equation} 
\label{theorem:main} 
\end{thm}

From problem definition, $P_Y(y_n)\!\!=\!\!\zeta_g(y_n)/||{\boldsymbol{\zeta}}_g||_1\!=\!1/N$ for all $n$. Substituting this uniform probability into Eq.~\ref{theorem:main}, the kernel can be further simplified as:
\begin{equation}
\begin{aligned}
    \boldsymbol{\mathcal{K}}_{i,n} 
    &= \frac{P(x_i|y_n) /N}{\sum_{n=1}^N P(x_i|y_n) /N} = \frac{P(x_i|y_n)}{\sum_{n=1}^N P(x_i|y_n)},\\
\end{aligned}
\label{eq:kernel_bay2}
\end{equation}
As suggested in Eq.~\ref{eq:kernel_bay2}, the kernel becomes fixed and pre-computable when the conditional distribution $P(x_i|y_n)$ is known or approximated (e.g., via a Gaussian assumption).

To define the loss function, we first discuss the transport cost term $W(\hat{\boldsymbol{P}})$. From Theorem~\ref{theorem:main}, $\hat{\boldsymbol{P}}$ is defined by a predeterminable $\boldsymbol{\mathcal{K}}$ and ${\boldsymbol{P}}_X$ reflecting input's ideal density. Since $\hat{\boldsymbol{P}}$ remains fixed during training, its cost is constant and does not contribute to model parameter optimization. Consequently, we effectively set the coefficient $\lambda_2=0$.


We focus solely on the transport discrepancy term $D\big(\tau(\tilde{{\boldsymbol{\zeta}}}_d^t), {\boldsymbol{\zeta}}_{g}\big)$. Theorem~\ref{theorem:main} ensures $\left\| \boldsymbol{\mathcal{K}}'{\boldsymbol{P}}_X \!\!-\!\!{\boldsymbol{P}}_Y\right\|_1\!=\!0$. 
For the ideal density distribution, where $\|{\boldsymbol{\zeta}}_d\|_1\!\!=\!\!\|{\boldsymbol{\zeta}}_g\|_1$, by scaling with their mass sums, $\left\| \boldsymbol{\mathcal{K}}'{\boldsymbol{\zeta}}_d-{\boldsymbol{\zeta}}_g\right\|_1\!=\!0$ holds. Consequently, for the estimated density distribution $\tilde{\boldsymbol{\zeta}_{d}^t}$, when the mass is proportionally pushed forward by $\boldsymbol{\mathcal{K}}$, the term
$\left\| \boldsymbol{\mathcal{K}}'\tilde{\boldsymbol{\zeta}_{d}^t}-\boldsymbol{\zeta}_{g} \right\|_1 \geqslant 0$ reflects the aligned mass discrepancy $D\big(\tau(\tilde{{\boldsymbol{\zeta}}}_d^t), {\boldsymbol{\zeta}}_{g}\big)$. Therefore, the Bayesian transport-based loss is defined as:
\begin{equation}
L_{BT} = \left\| \boldsymbol{\mathcal{K}}'\tilde{\boldsymbol{\zeta}_{d}^t}-\boldsymbol{\zeta}_{g} \right\|_1.
\label{eq:loss_bt}
\end{equation}

\begin{figure*}[t]
\centering
\includegraphics[width=0.85\linewidth]{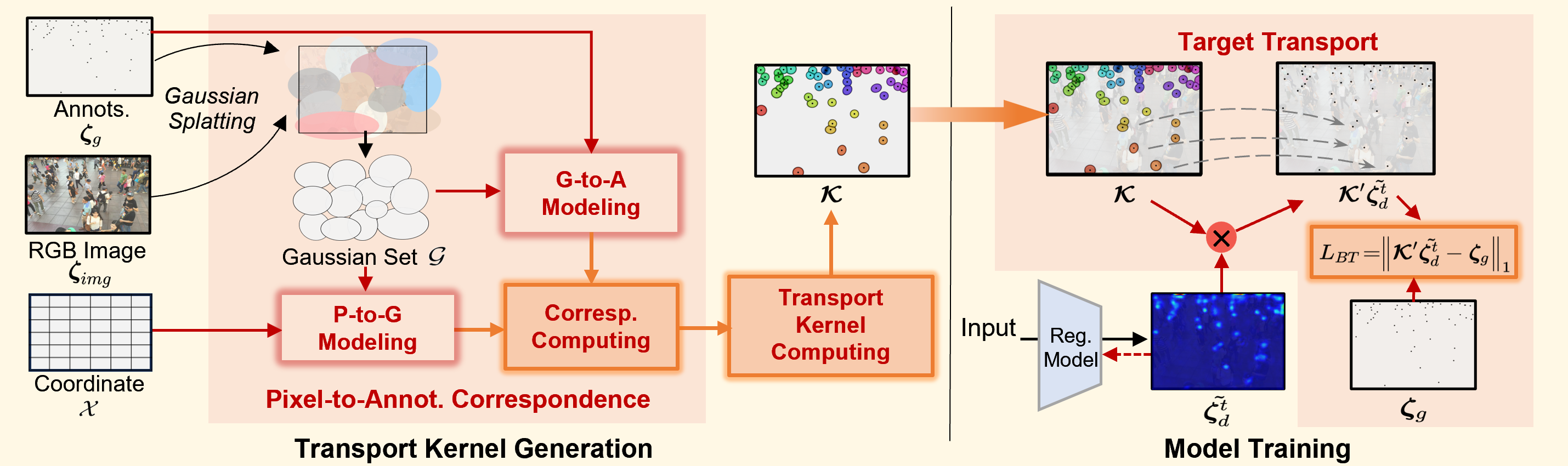}
\caption{The GST Pipeline comprises two main components: transport kernel generation and model training. First, the transport kernel $\boldsymbol{\mathcal{K}}$ is generated before training by reconstructing the RGB image via 2D Gaussian splatting and establishing pixel-to-annotation correspondences (Eq.~\ref{eq:P(x|y)}) to then form $\boldsymbol{\mathcal{K}}$ (Eq.~\ref{eq:kernel_bay2}). Second, during training, $\boldsymbol{\mathcal{K}}$ transports the estimated density map to annotations, allowing for the computation of the transported mass discrepancy loss (Eq.~\ref{eq:loss_bt}).}
\label{fig: pipeline}
\end{figure*}

As established, the efficacy of Bayesian Transport hinges on the pre-computation of the transport kernel $\boldsymbol{\mathcal{K}}$, which in turn depends on estimating the conditional probability 
$P(x_i|y_n)$. We introduce Gaussian Spatial Transport as our concrete realization of this framework. GST leverages 2D Gaussian Splatting to effectively model the spatial relationships within the image and derive a high-quality kernel. The following sections will detail the complete GST pipeline, from kernel generation to model training, as shown in Fig.~\ref{fig: pipeline}.

\subsection{Transport Kernel Generation via 2DGS}
\subsubsection{Pixel to annotation Correspondence}
\label{P2A}
To construct $\boldsymbol{\mathcal{K}}$, the core challenge lies in accurately approximating $P(x_i|y_n)$. We achieve this by using 2D Gaussian Splatting to explicitly model the underlying spatial correspondence between pixels and annotations, as its differentiable probabilistic representation is ideal for capturing the geometric features that govern this correspondence. The entire process is formally expressed using the law of total probability:
\begin{equation}
\begin{aligned}
    P(x_i|y_n)
    &= \sum\nolimits_{m=1}^M P(x_i|y_n, G_m)P(G_m|y_n) \\
    &= \sum\nolimits_{m=1}^M P(x_i|G_m)P(G_m|y_n),
\end{aligned}
\label{eq:P(x|y)}
\end{equation}
where the Gaussian ellipse $G_m \in \mathcal{G}$ is obtained via Gaussian Splatting and encodes the spatial distribution of the input image. The final equality in Eq.~\ref{eq:P(x|y)} holds due to the conditional independence of 
$x$ and $y$ given $G_m$. This effectively decomposes the original pixel-to-annotation correspondence problem into two manageable steps: computing the pixel-to-Gaussian correspondence $P(x_i|G_m)$ and the Gaussian-to-annotation correspondence $P(G_m|y_n)$. Next, we elaborate on how to compute these two components.

To establish \textbf{pixel-to-Gaussian} correspondences, we use Gaussian splatting~\cite{3dgs} to model the image as a composition of anisotropic Gaussian ellipses with fields of color and transparency. The image is approximated as:
\begin{equation}
    \begin{aligned}
        \tilde{\boldsymbol{\zeta}}_{\text{img}}(x_i) = \sum\nolimits_{m=1}^{M} \alpha_m\boldsymbol{c}_m \exp\left(-\frac{1}{2}d_M^2(x_i,G_m) \right), \\
        \ \ \text{where} \ \ d_M(x_i,G_m) = (x_i - \boldsymbol{\mu}_m)^\top \boldsymbol{\Sigma}_m^{-1} (x_i - \boldsymbol{\mu}_m),  \\ 
    \end{aligned}
\end{equation}
where each Gaussian component combines visual properties for rendering~(opacity $\alpha_m$ and color $\boldsymbol{c}_m$) with geometric parameters~(mean $\boldsymbol{\mu}_m$ and covariance $\boldsymbol{\Sigma}_m$) that determine its spatial distribution. 
$\boldsymbol{\Sigma}_m=\boldsymbol{R}_m\boldsymbol{S}_m(\boldsymbol{R}_m\boldsymbol{S}_m)'$ is parameterized by the scaling matrix $\boldsymbol{S}_m$ and rotation matrix $\boldsymbol{R}_m$, ensuring positive semi-definiteness during gradient-based optimization, as in~\cite{3dgs}. 
The pixel-to-Gaussian correspondence is derived \emph{exclusively} from the spatial probability distribution:
\begin{equation}
    \begin{aligned}
        P(x_i|G_m) = \mathcal{N}(x_i;\boldsymbol{\boldsymbol{\mu}}_m,\boldsymbol{\Sigma}_m), \ \  G_m \in \mathcal{G}
    \end{aligned}
    \label{eq:P(x|G)}
\end{equation}

\noindent\textit{Deformity Elimination.}
To ensure Gaussians match annotated objects, we introduce a shape control regularization to maintain reasonable shapes:
\begin{equation}
L_{shape} = \max_{m}(\max(s_{ml}/{s_{ms}}-\delta,0)),
\label{eq:shape loss}
\end{equation}
where $s_{ml}$,$s_{ms}$ denote the scales along the major and minor axes of the Gaussian ellipse, respectively, and $\delta=1.5$ controls the maximum permissible aspect ratio. This loss term prevents unconstrained Gaussians from forming excessively elongated deformations that could erroneously focus on irrelevant narrow regions, thereby maintaining accurate spatial correspondence between pixels and annotations. The total loss for Gaussian splatting process becomes $L_{gs} = L_{rec} + \beta L_{shape}$\footnote{$L_{gs}$ is specifically used to optimize the parameters of the Gaussians, rather than those of the task-tailored model $f$.}, where $L_{rec}$ is the reconstruction loss~\cite{gaussianimage} implemented by mean square distance between $\tilde{\boldsymbol{\zeta}}_{img}$ and ${\boldsymbol{\zeta}}_{\text{img}}$.

With Gaussians generated, we model the \textbf{Gaussian-to-Annotation} correspondence by a similarity metric, i.e., $P(G_m|y_n) = \frac{\exp(\text{sim}(G_m, y_n))}{\sum_{m} \exp(\text{sim}(G_{m}, y_n))}$. While it can be computed by evaluating the Gaussian probability density function at the annotation point, the pairwise computation of annotation-Gaussian similarities becomes computationally exhaustive and memory-intensive when dealing with large numbers of Gaussians and annotations. Moreover, in dense scenarios (e.g., in the task of crowd-counting), the image region corresponding to an annotation point is typically small and approximately elliptical.

Based on above analysis, we simplify the correspondence modeling by establishing a one-to-one correspondence between annotations and Gaussians through a pre-assignment strategy within the Gaussian splatting process. Specifically, we partition a foreground Gaussian set $\mathcal{G}_{fg} \subseteq \mathcal{G}$, where each Gaussian $G_m \in \mathcal{G}_{fg}$ is uniquely pre-assigned to a specific annotation through an assignment mapping $\mathrm{asgn}(m)$. These Gaussians are centred at the position of the assigned annotation throughout the parameter optimization of GS. To maintain strict one-to-one correspondence, we ensure $|\mathcal{G}|\geq N$ and $|\mathcal{G}_{fg}|=N$, with the remaining Gaussians randomly initialized to model unannotated image regions. This configuration creates \emph{binary correspondences} where each foreground Gaussian exclusively corresponds to its pre-assigned annotation with probability 1 while having zero correspondence with all other annotations.

\subsubsection{Pixel to Background Correspondence}

Given the sparsity of foreground targets and the significant proportion of background pixels in an image, OT's approach of assigning all pixels to targets is suboptimal for background pixels.
To address this, we augment the annotation set $\mathcal{Y}$ with a virtual background object and develop an auxiliary mechanism for background correspondence. Specifically, pixels are assigned to the background if their distance to the nearest foreground Gaussian exceeds a predefined threshold. 
Furthermore, for consistency, we extend the previously defined $\mathcal{G}, \mathcal{Y}$ by introducing a background pseudo-Gaussian $G_0$ and a virtual background object $y_0$, i.e., $\mathcal{G} \leftarrow \mathcal{G} \cup \{G_0\}$, $\mathcal{Y} \leftarrow \mathcal{Y} \cup \{y_0\}$, with $\boldsymbol{\zeta}_g(y_0)=0$ to prevent background mass allocation. 
The correspondence of pixel $x_i$ given background region $G_0$ is defined as:
\begin{equation}
\begin{aligned}
    P({x}_i|G_0)
    &\!\triangleq\!
    \frac{1}{2\pi|\boldsymbol{\Sigma}_*|^{1/2}}\exp\!\left(\!-\frac{d^2 \! - \! d_{M}^{2}{\left( {x}_i, G_*\!\right)}}{2}\!\right), \\
    &\text{where} \ \ G_*\!= \!\arg\min_{G \in {\mathcal{G}_{fg}}} \! d_{M} \!\left( {x}_i, G\right)\!.
\end{aligned}
\label{eq:P(x|Bk)}
\end{equation}
$G_*$ represents the nearest foreground Gaussian to $x_i$, determined by the Mahalanobis distance $d_{M}{\left(\! \boldsymbol{x}_i, G\!\right)}$. $\boldsymbol{\Sigma}_*$ denotes its covariance. $d$ serves as a cut-off threshold to control the spatial extent of background region assignment.

The correspondence between $G_m$ and target $y_n$ for $m, n  \geq 0$ is determined by the pre-assigned mapping:
\begin{equation}
P(G_m|y_n) = 
      \mathbb{I}(\mathrm{asgn}(m),y_n)
  \label{eq:P(G|y)_bk}
\end{equation}
where $\mathbb{I}$ is an indicator function returning 1 when inputs are equal. 
Integrating background terms from Eqs.~\ref{eq:P(x|Bk)} and \ref{eq:P(G|y)_bk} into Eq.~\ref{eq:P(x|y)} easily yields the complete pixel-to-background correspondence.

\subsection{Model Training and Optimization}
With the transport kernel $\mathcal{K}$ generated as described, the model training proceeds as the final phase of our pipeline, as in Fig.\ref{fig: pipeline}. The optimization is driven by the loss previously defined in Eq.~\ref{eq:loss_bt}, , which leverages the pre-computed $\mathcal{K}$.
In contrast to iterative OT-based methods, our loss is calculated via a single, efficient matrix multiplication. This design makes the training process highly efficient and directly optimizes the model to produce density maps that align with the rich spatial correspondences encoded in the kernel.

\section{Experiment Results and Discussion}

\label{sec:experiment}
We evaluate the proposed Gaussian Spatial Transport~(GST) on crowd counting and landmark localization, representative tasks for point-supervised density regression. For crowd counting, we use three benchmarks: UCF-QNRF~\cite{ucfqnrf}, JHU-Crowd++~\cite{sindagi2020jhu}~\cite{nwpu}, and NWPU, reporting MAE and MSE. For landmark localization, we use the MPII Human Pose benchmark~\cite{simonyan2014vgg}, reporting PCKh@0.5. 2DGS is implemented using gsplat~\cite{ye2024gsplatopensourcelibrarygaussian}, with the cut-off distance $d$ set to 3. Further implementation details are in \emph{Suppl. C}.

\subsection{Evaluation Results}
\begin{figure}[ht]
\centering
\includegraphics[width=1.0\linewidth]{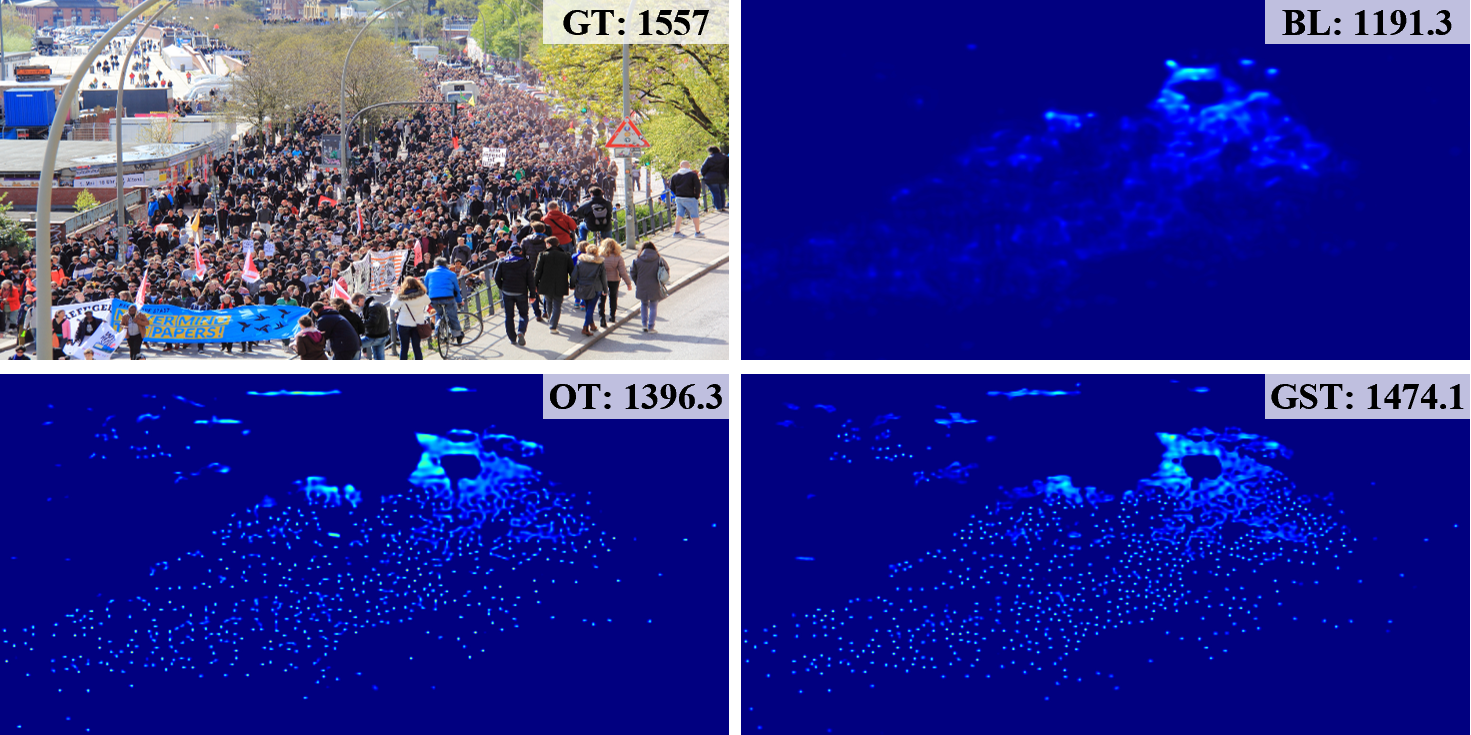}
\caption{Visualization of crowd counting.}
\label{fig:counting}
\end{figure}

\paragraph{Crowd Counting.}
Table~\ref{tab:counting comparison1} focuses on the datasets UCF-QNRF~\cite{ucfqnrf} and JHU-Crowd++~\cite{sindagi2020jhu}. We notably excel the $L_2$ Baseline~(direct regression) and handcrafted fixed transport plan~(BL). On JHU-Crowd++, we achieve the lowest MAE~53.9 and MSE~225.4, marginally outperforming APGCC. Compared to methods based on optimal transport~(DMCount, UOT, GL), our approach achieved significant performance improvements On UCF-QNRF, reducing MSE by 17.2 against balanced OT~(DMCount), and 11.2 and 16.4 against unbalanced OT~(UOT and GL, respectively). Fig.~\ref{fig:counting} visualizes the estimated density map of BL, OT~(DMCount), and GST. Besides accurate counting estimation, GST also obtains \emph{sharp} density distributions as the OT-based method. 

Table~\ref{tab:counting comparison2} lists the results on NWPU~\cite{nwpu}, which is currently the largest, most diverse, and most challenging crowd counting dataset. We consistently achieves strong performance, highlighting the efficiency of our transport strategy, particularly advantageous for tackling high-density scenarios and complex environmental conditions. 

\begin{table}[th]
\setlength{\tabcolsep}{1mm}
\footnotesize
	\centering
		\begin{tabular}{c|cc|cc}
			\toprule
			\multicolumn{1}{c}{Dataset}
            & \multicolumn{2}{c}{JHU++} & \multicolumn{2}{c}{UCF-QNRF}\\ 
			
			\multicolumn{1}{c}{Method} &
                \multicolumn{1}{c}{MAE} & \multicolumn{1}{c}{MSE} &
                \multicolumn{1}{c}{MAE} & \multicolumn{1}{c}{MSE} \\
			\hline
            $L_2$ Baseline
            & 81.7 & 304.5 & 107.2 & 164.6 \\ 
            BL~\cite{ma2019bayesian}
            & 75.0 & 299.9 & 88.7 & 154.8 \\ 
            DMCount~\cite{wang2020dm}
            & 61.6 & 256.1 & 85.6 & 148.3 \\ 
            UOT~\cite{ma2021learning}
            & 60.5 & 252.7 & 83.3 & 142.3 \\ 
            GL~\cite{wan2021generalized}
            & 59.5 & 259.5 & 84.3 & 147.5 \\
            P2PNet~\cite{song2021P2P}
            & -- & -- & 85.3 & 154.5 \\ 
            \cite{liang2022CLTR} 
            & 59.5 & 240.6 & 85.8 & 141.3 \\ 
            PET~\cite{liu2023PET}
            & 58.5 & 238 & \textbf{79.5} & 144.3 \\ 
            APGCC~\cite{chen2024APGCC}
            & 54.3 & 225.9 & 80.1 & 136.6 \\ 
            \rowcolor{gray!15}
            Ours
            & \textbf{53.9} & \textbf{225.4} & 80.7 & \textbf{131.1} \\ 
		\bottomrule
		\end{tabular}
\caption{Counting Performance on JHU++ and UCF-QNRF.} 
\label{tab:counting comparison1}
\end{table}

\begin{table}[th]
\footnotesize
	\centering
		\begin{tabular}{c|cc}
			\toprule
			\multicolumn{1}{c}{Method} & \multicolumn{1}{c}{MAE} & \multicolumn{1}{c}{MSE}\\
			\hline
            $L_2$ Baseline & 126.2 & 528.2 \\
            BL~\cite{ma2019bayesian} 
            & 105.4 & 454.2 \\
            DMCount~\cite{wang2020dm}
            & 88.4 & 388.6 \\
            UOT~\cite{ma2021learning}
            & 87.8 & 387.5 \\
            GL~\cite{wan2021generalized}
            & 79.3 & 346.1 \\
            MAN~\cite{lin2022boosting}
            & 76.5 & 323.0 \\
            ChfL~\cite{shu2022chfl}
            &76.8	&343\\
            PET~\cite{liu2023PET}
            &74.4	&328.5\\
            \rowcolor{gray!15}
            Ours
            &\textbf{74.4} & \textbf{306.2} \\
		\bottomrule
		\end{tabular}
\caption{Performance of crowd counting on NWPU.}
\label{tab:counting comparison2}
\end{table}

\begin{figure}[t]
\centering
\includegraphics[width=1.0\linewidth]{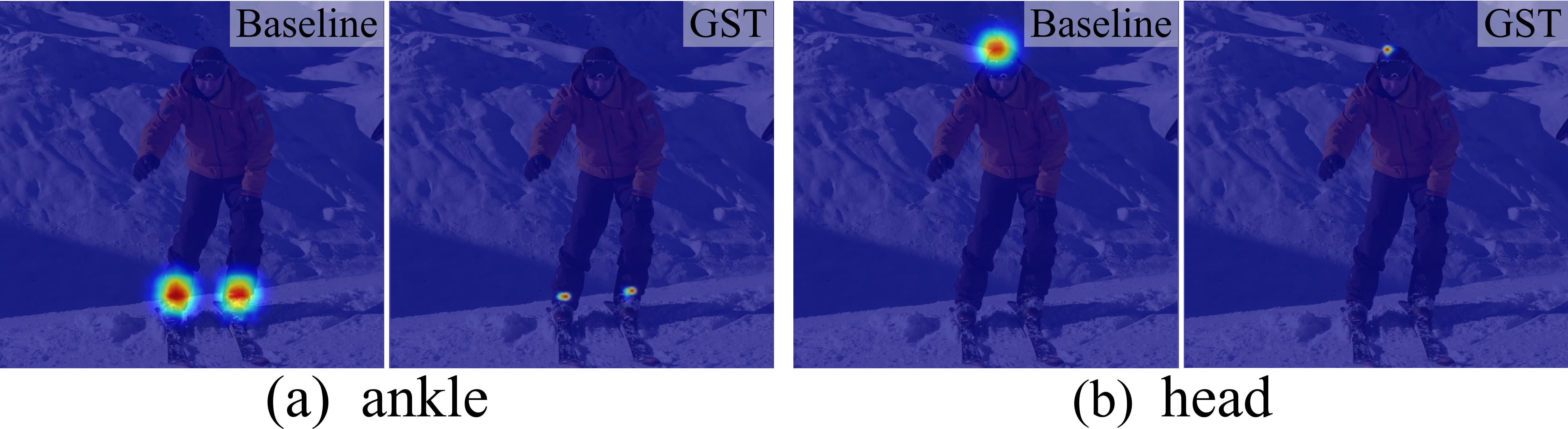}
\caption{Visualization of landmark location.}
\label{fig:landmark}
\end{figure}

\paragraph{Landmark Localization.}
We evaluate our transport strategy against two approaches: Gaussian-blurred pseudo-density map regression~(HRNet) and optimal transport cost minimization~(HDM-HPE). 
As in Table~\ref{Tab:mpii}, our method achieves the best mean PCKh@0.5 across all joints, outperforming the baseline HRNets by 0.6\% on HRNet-W32/W48 architectures, with over 1.0\% gains on challenging joints like shoulders, hips, knees, and ankles\footnote{The observed performance discrepancy in \emph{head} stems from annotation bias: MPII's head joint annotations are at the periphery, not the center, degrading performance.}. Comparable to OT-based HDM-HPE, our approach directly uses raw point annotations, avoiding computationally intensive subpixel smoothing~\cite{qu2022OTHPE}. Fig.~\ref{fig:landmark} shows that GST generates sharper density peaks, clarifying the location.

\begin{table}[ht]
\footnotesize
\centering
\setlength{\tabcolsep}{1mm}
\label{Tab:mpii_transposed}
\begin{tabular}{l|ccc|ccc} 
\toprule
& \multicolumn{3}{c|}{HRNet-W32} & \multicolumn{3}{c}{HRNet-W48} \\
Metric & HRNet & HDM-HPE & Ours & HRNet & HDM-HPE & Ours \\
\hline
\hline
Mean & 90.4 & 90.9 & \textbf{91.0} & 90.5 & 90.9 & \textbf{91.1} \\
\hline
Head & 97.1 & 97.3 & 96.2 & 96.9 & 97.1 & 96.4 \\
Shldr. & 95.9 & 96.2 & 97.1 & 96.0 & 96.3 & 97.0 \\
Elb. & 90.7 & 91.2 & 90.3 & 90.9 & 91.2 & 90.9 \\
Wri. & 86.1 & 86.8 & 86.9 & 86.2 & 87.0 & 86.5 \\
Hip & 89.4 & 90.1 & 90.7 & 89.6 & 90.2 & 91.1 \\
Kne. & 86.9 & 87.4 & 88.4 & 87.1 & 87.5 & 88.0 \\
Ank. & 83.2 & 84.1 & 85.1 & 83.5 & 84.2 & 84.7 \\
\bottomrule
\end{tabular}
\caption{Performance of human pose estimation on MPII.}
\label{Tab:mpii}
\end{table}

\subsection{Contribution of Components and Parameters}

\paragraph{Ablation study.}
Table~\ref{tab:ablation} reports the ablation study results organized on a VGG~\cite{simonyan2014vgg} backbone. The heuristic transport plan, which computes the transport kernel $\mathcal{K}$ via a Gaussian $P(x_i|y_n)$ with an empirical $\sigma=8$, significantly outperforms the $L_2$ baseline, highlighting the advantage of Bayesian transport modeling. Furthermore, the 2DGS transport plan, derived from our proposed method leveraging Gaussian Splatting, demonstrates clear superiority over the heuristic plan. Finally, the inclusion of pixel-to-background correspondence further enhances performance, confirming its effectiveness in suppressing irrelevant background regions.

Meanwhile, we conducted experiments on VGG and Transformer architectures~(as in~\cite{lin2024gramformer}) to further compare the performance of the OT method and our GST. The results in Table~\ref{tab:counting arch.} show that GST surpasses OT on both VGG and Transformer, highlighting its consistent robustness regardless of the backbone architecture.

\begin{table}[thbp!]
\footnotesize
\center
\begin{tabular}{ccc|cc}
\toprule[1pt]
transport plan & P2A Cor. & P2B Cor. & MAE & MSE \\
\hline
$L_2$ baseline &  & & 81.70 & 304.50  \\
heuristic plan & & & 65.45 & 268.50\\
GST- & \checkmark & & 60.17 & 247.61 \\
GST & \checkmark &\checkmark & \textbf{58.30} & \textbf{239.58} \\
\toprule[1pt]
\end{tabular}
\caption{Ablation Studies on JHU++ with VGG backbone. 
\emph{P2A Cor.} and \emph{P2B Cor.} represents the Pixel-to-Annotation and Pixel-to-background Correspondence, respectively.}
\label{tab:ablation}
\end{table}

\begin{table}[th]
\setlength{\tabcolsep}{1mm}
\footnotesize
        \centering
		\begin{tabular}{c|c|cc|cc|cc}
			\toprule
			\multirow{2}{*}{Arch.} & \multirow{2}{*}{Method} & \multicolumn{2}{c}{UCF-QNRF} &  \multicolumn{2}{c}{JHU++} & \multicolumn{2}{c}{NWPU} \\
			
                & \ & \multicolumn{1}{c}{MAE} & \multicolumn{1}{c}{MSE} &  
                    \multicolumn{1}{c}{MAE} & \multicolumn{1}{c}{MSE} &  
                    \multicolumn{1}{c}{MAE} & \multicolumn{1}{c}{MSE}\\
                \hline
                \multirow{2}{*}{VGG} & OT & 87.2 & 150.8 & 61.6 & 256.1 & 88.4 & 388.6 \\
                                     & GST &\textbf{82.5} & \textbf{139.2} & \textbf{58.3} & \textbf{239.6} & \textbf{80.3} & \textbf{325.7} \\
                \hline
                \multirow{2}{*}{Trans.} & OT & 87.8 & 153.0 & 58.5 & 240.0 & 80.9 & 323.5 \\
                                       & GST & \textbf{80.7} & \textbf{131.1} & \textbf{53.9} & \textbf{225.4} & \textbf{74.4} & \textbf{306.2} \\
		\bottomrule
		\end{tabular}
\caption{Counting Results on different model architectures. \emph{Trans} is the abbreviation of Transformer.}
\label{tab:counting arch.}
\end{table}

\begin{figure}[ht]
\centering
\includegraphics[width=1.0\linewidth]{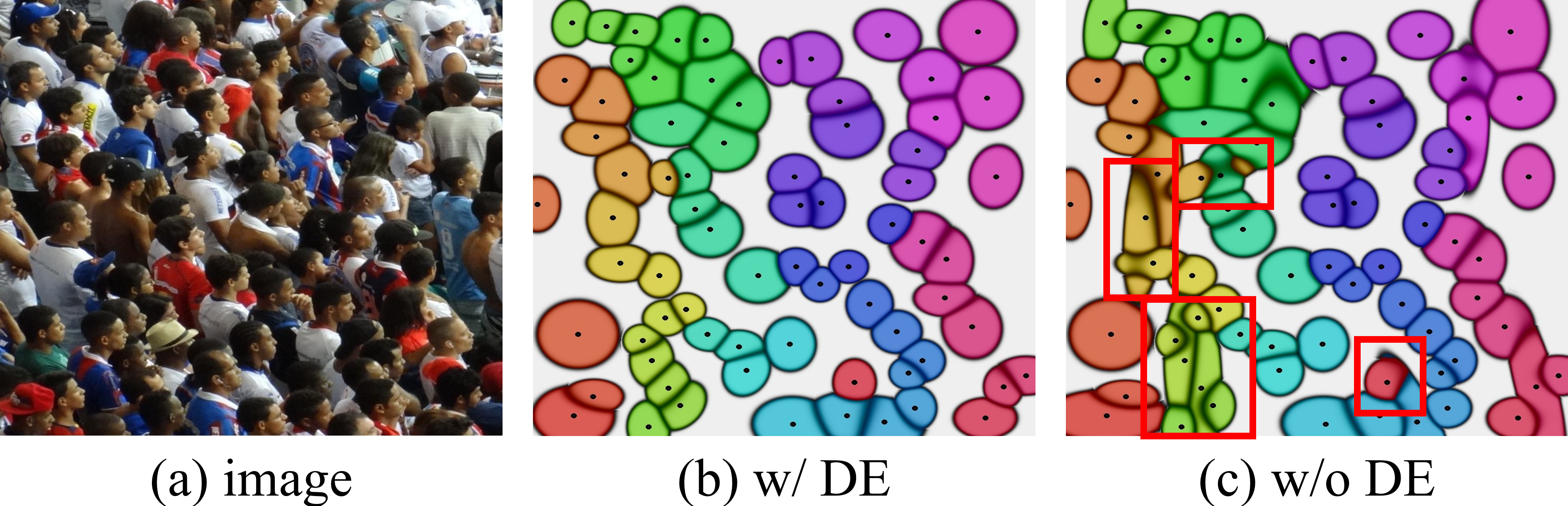}
\caption{Visualization of transport plan w/ and w/o deformity elimination~(DE) during Gaussian splatting. The map shows annotation-to-pixel correspondence: black points are annotations, unique colors represent individual targets with brightness indicating transport strength, and blank regions denote background transport.}
\label{fig:transport}
\end{figure}

\paragraph{Effect of Deformity Elimination.}
We investigate the necessity of deformity elimination in the splatting process in Fig.~\ref{fig:transport}.
Without it, over-elongated Gaussians erroneously cover irrelevant background regions, causing significant overlapping transport regions (Fig.~\ref{fig:transport}(c), red boxes). This leads to penetrating-object transport and ambiguous correspondences, resulting in unreliable training supervision.

\begin{figure}[ht]
\centering
\includegraphics[width=1.0\linewidth]{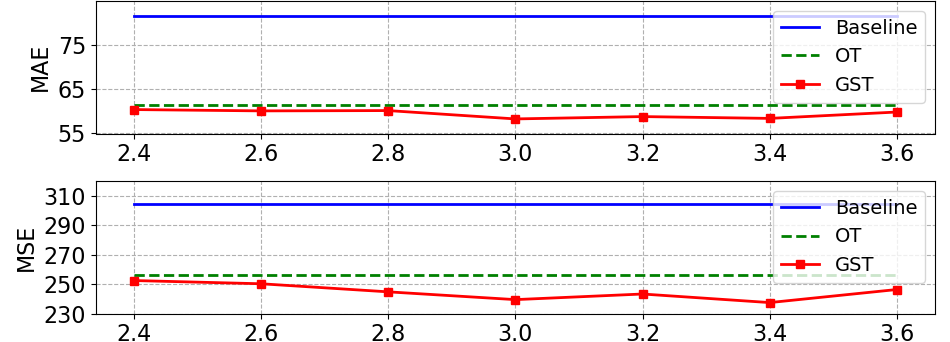}
\caption{Results w.r.t. $d$ on JHU++ with VGG backbone.}
\label{fig:cut_off}
\end{figure}

\paragraph{Effect of $d$ in Pixel-to-Background Correspondence.}
We conduct experiments to evaluate the effect of cut-off distance $d$ on VGG. As in Fig~\ref{fig:cut_off}, the optimal performance is achieved at $d=3$,  which aligns with the $3\sigma$ principle in 2DGS rendering, ensuring effective foreground-background separation.  Moreover, our method is robust to parameter selection near this optimum, consistently outperforming OT within the $2.4\sim3.6$ range.

\subsection{Discussion}
\label{sec:discussion}

\paragraph{The role of 2D Gaussian Splatting.}

Gaussian Splatting plays a central role in building a spatial transport plan from density maps to annotations as in Eqs.~\ref{eq:kernel_bay2} and \ref{eq:P(x|y)}. Fig.~\ref{fig:transport} exemplifies a plan. It can be observed that GST establishes a discrete spatial mapping from image coordinate space to annotation space. 
To further quantify the effectiveness of GST, we compare it against a transport plan using \emph{ideal} ground truth scale information, alongside heuristic-based scales and the OT plan, as shown in Table~\ref{tab: scales}. Transport plans based on ground truth scales and 2DGS-estimated scales outperform both the OT plan and the heuristic-based plan.

These evaluations demonstrate that GST effectively estimates head scales using only point annotations. Moreover, its transport plan outperforms both the OT-based and the static heuristic-based approach. Finally, the results validate our assumption that a reliable transport plan can be derived from the input image and annotated points prior to task-specific network optimization.

\paragraph{Differences between OT and GST.} 
GST is fundamentally different from traditional OT, representing \emph{a distinct paradigm rather than an approximation}. The foundational difference lies in the source of information used for planning: while OT-based approaches must wait for and operate on the network's density estimation, our method directly constructs a correspondence from the input RGB image itself. This distinction in methodology leads to two key advantages.
First, the RGB input space provides richer semantic information than just a one-dimensional density map. This allows GST to generate more interpretable transport plans that are grounded in the actual image content.
Second, and more critically, GST's approach decouples the costly planning from the iterative optimization process. This is achieved through a one-time pre-computation of the transport kernel. To quantify this efficiency gain, consider an image with $p$ annotated points and $q$ pixels. 
The pre-computation is dominated by computing the probability density function between coordinates and kernels, costing $\mathcal{O}(pq)$. Crucially, this computation occurs before training, reducing the theoretical complexity \emph{during training} to $\mathcal{O}(1)$.
This complete separation of planning from training avoids the repeated calculations inherent to OT and creates a win-win solution, free from the trade-offs typical of two-stage optimization. In stark contrast, OT requires a much higher complexity of 
$\mathcal{O}(kpq)$ for iterative Sinkhorn optimization within each training step, typically $k$=100 for DMCount~\cite{wang2020dm} and $k$=1,000 for HDM-HPE~\cite{qu2022OTHPE}. This theoretical difference translates directly to practice, as GST reduces the runtime of OT almost by half on an NVIDIA 4090, as shown in the last column of Table~\ref{tab: scales}.

As GST involves a pre-computation step, its cost is minimal. This one-time process typically takes only 3-7 seconds per high-resolution image, and remains under 10 seconds even for extreme cases with resolutions up to 4K and over 4,500 annotated points. 
This runtime could be further reduced with optimized Gaussian Splatting implementations. Consequently, the overall pipeline is significantly faster than end-to-end OT-based training. More discussion on differences between OT and GST can be referred to~\emph{Suppl. D}.

\begin{table}[ht]
\footnotesize
\centering
\begin{tabular}{c|cc>{\columncolor{gray!15}}c}
\toprule[1pt]
Method &  MAE & MSE & Training Time  \\
\hline
plan with \emph{g.t.} scale & \textbf{56.60} & \textbf{238.24} & 15h29min \\
Heuristic plan & 65.45 & 268.50 & 14h35min \\
OT & 61.55 & 256.10 & 28h36min \\
GST & \underline{58.30} & \underline{239.58} & 15h32min \\
\toprule[1pt]
\end{tabular}
\caption{Accuracy and runtime comparison of different transport plans on JHU++ with VGG backbone.}
\label{tab: scales}
\end{table}

\section{Conclusion and Limitations}
\label{sec:conclusion}

This paper proposes Gaussian Spatial Transport, which leverages Gaussian splatting to compute the transport plan. We apply it to representative tasks, including crowd counting and landmark detection. Experimental results demonstrate its clear advantages over traditional optimal transport schemes in both training efficiency and accuracy. Moreover, GST successfully extends the application of Gaussian splatting in computer vision. Validating it in a broader range of applications presents a promising research direction.

However, our method still has certain limitations. For instance, the hard binding of the Gaussian kernels to the annotations during splatting optimization may be suboptimal. More flexible soft-assignment mechanisms can be further explored in future work. Additionally, the 2D Gaussian splatting based on a single image lacks explicit depth modeling, which is crucial for image understanding. Considering recent progress in 3D Gaussian splatting for sparse-view reconstruction, future extensions could incorporate 3D Gaussians with explicit depth information for the transport in the multi-view scene to enhance spatial reasoning capabilities.

\clearpage
\section{Acknowledgements}
\label{sec:ack}

This work was funded in part by the National Natural Science Foundation of China (62376070, 62076195) and in part by the Fundamental Research Funds for the Central Universities (AUGA5710011522). Thanks to all anonymous reviewers and the area chair for their valuable comments. Thanks to Mr. Yupeng Wei for his discussion on the optimal transport theory.

\bibliography{aaai2026}

\end{document}


\maketitle

\appendix 

This document provides supplementary information to accompany our main manuscript. Section~\ref{sec:supp_discrepancy} begins with a detailed discussion on the implementation of the discrepancy term $D(\cdot)$ in prior OT-based works. This is followed by a rigorous proof of our core theoretical result, Theorem 1, in Section~\ref{sec:supp_proof1}. To ensure reproducibility, comprehensive implementation details for all experiments, including pre-computation costs, are provided in Section~\ref{sec:supp_implementation}. Finally, Section~\ref{sec:supp_comparison} offers an in-depth comparison between our GST framework and traditional OT, highlighting their key conceptual differences.

\section{Discussion on the Discrepancy Term $D(\cdot)$}
\label{sec:supp_discrepancy}

In the main \emph{manuscript}, we introduced a general form for transport-based loss functions: 
\begin{equation}
L(\tilde{\boldsymbol{\zeta}}_{d}^t, \boldsymbol{\zeta}_{g}) = \lambda_1 D\big(\tau(\tilde{\boldsymbol{\zeta}}_{d}^t), \boldsymbol{\zeta}_{g}\big) + \lambda_2 W(\boldsymbol{P}). 
\end{equation}
Here, we provide a more detailed discussion on how the transport discrepancy term, $D(\cdot)$, is implemented in representative Optimal Transport (OT) based methods for crowd counting, namely DM-Count~\cite{wang2020dm}, UOT~\cite{ma2021learning}, and GL~\cite{wan2021generalized}. These methods differ primarily in how they handle the mass conservation constraint between the predicted density $\tilde{\boldsymbol{\zeta}}_{d}$ and the ground-truth annotations $\boldsymbol{\zeta}_{g}$ and therefore define this discrepancy term differently.

\subsection{DMCount: Explicit Count Discrepancy}
DMCount~\cite{wang2020dm} operates under the framework of \textbf{balanced} OT, which enforces a strict mass equality constraint. The transport cost $ W(\boldsymbol{P}_*^t)$ is therefore computed between the normalized distributions $\tilde{\boldsymbol{P}_X^t}$ and $\boldsymbol{P}_Y$. The balanced OT means the transport plan $\boldsymbol{P_*^t}$ must satisfy both marginal constraints perfectly, i.e., $\boldsymbol{P}_*^t\mathbf{1_N} = \tilde{\boldsymbol{P}_X^t}$ and ${\boldsymbol{P}_*^t}'\mathbf{1_I} = \boldsymbol{P}_Y$, where $\tilde{\boldsymbol{P}_X^t}$ and $\boldsymbol{P}_Y$ are normalized distributions. As a result, when transporting the normalized prediction $\tilde{\boldsymbol{P_X^t}}$, the discrepancy is zero by definition.

However, the predicted count (total mass) $\|\tilde{\boldsymbol{\zeta}}_{d}^t\|_1$ may not equal the ground-truth count $\|\boldsymbol{\zeta}_{g}\|_1$. Within our general framework, the discrepancy term for DM-Count is defined as the absolute difference in total mass:
\begin{equation}
D_{\text{DM}}(\tau(\tilde{\boldsymbol{\zeta}}_{d}^t), \boldsymbol{\zeta}_{g}) = \big| \|\tilde{\boldsymbol{\zeta}}_{d}^t\|_1 - \|\boldsymbol{\zeta}_{g}\|_1 \big|.
\end{equation}
As noted in the main \emph{Manuscript}, this is equivalent to the $L_1$ norm of the difference between the pushed-forward unnormalized distribution~(proportionally by the plan $\boldsymbol{P}_*^t$) and the ground truth, i.e., $||\tau(\tilde{\boldsymbol{\zeta}}_{d}^t) - \boldsymbol{\zeta}_{g}||_1$. 

The proof of this equivalence is as follows. 
The push-forward operator $\tau(\cdot)$ transports  $\tilde{\boldsymbol{\zeta}}_{d}^t$ according to the optimal plan $\boldsymbol{P}_*^t$, which was computed on normalized distributions. The resulting transported mass at the target locations is the target marginal of this unnormalized transport.

\begin{framedproof}{Proof of Equivalence \\ $||\tau(\tilde{\boldsymbol{\zeta}}_{d}^t) - \boldsymbol{\zeta}_{g}||_1=\big| \|\tilde{\boldsymbol{\zeta}}_{d}^t\|_1 - \|\boldsymbol{\zeta}_{g}\|_1 \big|$}

\begin{align}
    &\big\|\tau(\tilde{\boldsymbol{\zeta}}_{d}^t) - \boldsymbol{\zeta}_{g}\big\|_1\\
    &= \big\| ( \|\tilde{\boldsymbol{\zeta}}_{d}^t \|_1 \cdot \boldsymbol{P}_*^t ) ' \mathbf{1}_I - \boldsymbol{\zeta}_{g} \big\|_1 \tag{by definition of $\tau$} \label{step:def_tau} \\
    &= \big\| \|\tilde{\boldsymbol{\zeta}}_{d}^t \|_1 \cdot ( {\boldsymbol{P}_*^t} ' \mathbf{1}_I) - \boldsymbol{\zeta}_{g} \big\|_1 \tag{linearity of matrix operations} \label{step:linearity} \\
    &= \left\| \|\tilde{\boldsymbol{\zeta}}_{d}^t \|_1 \cdot \frac{\boldsymbol{\zeta}_{g}}{\| \boldsymbol{\zeta}_{g} \|_1} - \boldsymbol{\zeta}_{g} \right\|_1 
    \tag{since ${\boldsymbol{P}_*^t}' \mathbf{1}_I = \boldsymbol{P}_Y = \frac{\boldsymbol{\zeta}_{g}}{\|\boldsymbol{\zeta}_{g}\|_1}$} \label{step:marginal_constraint} \\
    &= \left\| \left( \frac{\|\tilde{\boldsymbol{\zeta}}_{d}^t \|_1}{\| \boldsymbol{\zeta}_{g} \|_1} - 1 \right) \boldsymbol{\zeta}_{g} \right\|_1 \tag{factoring out $\boldsymbol{\zeta}_{g}$} \label{step:factoring} \\
    &= \left| \frac{\|\tilde{\boldsymbol{\zeta}}_{d}^t \|_1}{\| \boldsymbol{\zeta}_{g} \|_1} - 1 \right| \cdot \|\boldsymbol{\zeta}_{g}\|_1 \tag{property of $L_1$ norm: $\|c \cdot \mathbf{v}\|_1 = |c| \cdot \|\mathbf{v}\|_1$} \label{step:norm_property} \\
    &= \left| \|\tilde{\boldsymbol{\zeta}}_{d}^t \|_1 - \| \boldsymbol{\zeta}_{g} \|_1 \right| \tag{simplifying} \label{step:final}
\end{align}
This concludes the proof.

\end{framedproof}

The final loss in DM-Count is thus a weighted sum of this discrepancy term and the Wasserstein distance, $L_{DM} = \lambda_D D_{\text{DM}}(\tau(\tilde{\boldsymbol{\zeta}}_{d}^t), \boldsymbol{\zeta}_{g}) + W(\boldsymbol{P}_*^t)$.

\subsection{UOT: Implicit Discrepancy via KL Penalties}
Unbalanced Optimal Transport (UOT), as implemented in Ma et al.~\cite{ma2021learning}, relaxes the strict marginal constraints to better handle cases where the predicted and ground-truth masses naturally differ. It fundamentally differs from balanced OT by directly handling measures with unequal total mass, i.e., unnormalized measures $\tilde{\boldsymbol{\zeta}}_{d}^t$ and $\boldsymbol{\zeta}_g$, making it naturally suited for crowd counting, where the predicted count $\|\tilde{\boldsymbol{\zeta}}_{d}^t \|_1$ may differ from the ground-truth count $\|\boldsymbol{\zeta}_{g} \|_1$. 

In the context of our general loss form, the UOT formulation effectively embeds the discrepancy term $D(\cdot)$ into the transport problem itself. The concept of a separate discrepancy term $D(\cdot)$ is thus \textbf{implicit}. The entire loss is the solution to a single, unified optimization problem that jointly minimizes transport cost and marginal deviations:
\begin{equation}
\begin{aligned}
    L_\text{UOT} = &\min_{\boldsymbol{P} \in \mathbb{R}_+^{I \times N}} \big( \langle \mathbf{C}, \boldsymbol{P} \rangle + \lambda_{kl} \text{KL}(\boldsymbol{P}\mathbf{1}_N || \tilde{\boldsymbol{\zeta}}_{d}^t) \\
    & + \lambda_{kl} \text{KL}(\boldsymbol{P}'\mathbf{1}_I || \boldsymbol{\zeta}_{g}) \big)
\end{aligned}
\end{equation}
Here, the KL terms measure the "discrepancy" between the marginals of the transport plan and the desired source/target distributions. $\text{KL}(\boldsymbol{P}\mathbf{1}_N || \tilde{\boldsymbol{\zeta}}_{d}^t)$ measures the divergence between the total mass transported from each pixel and the predicted density value at that pixel. $\lambda_{kl} \text{KL}(\boldsymbol{P}'\mathbf{1}_I || \boldsymbol{\zeta}_{g})$ measures the divergence between the total mass transported to each annotation and the ground-truth count at that annotation, which is typically 1. The hyperparameter $\lambda_{kl}$ controls the "softness" of the marginal constraints. A larger $\lambda_{kl}$ pushes the solution closer to balanced OT. 

In this case, one could conceptually define the "discrepancy term" as the two KL penalties, but they are not computed separately from the transport cost. The optimization problem is solved as a whole.

\subsection{GL: A Hybrid Approach}
Generalized Loss (GL)~\cite{wan2021generalized} introduces a sophisticated hybrid framework that attempts to capture the benefits of both total count accuracy and localization-aware transport. It explicitly separates the pixel-wise and point-wise discrepancies from the transport cost, similar to our general form. The key distinction from other methods lies in its choice of discrepancy measures. For GL, the discrepancy term $D(\cdot)$is \textbf{an explicit sum of two different norms}:
\begin{equation}
D_{\text{GL}}(\tau(\tilde{\boldsymbol{\zeta}}_{d}^t),\boldsymbol{\zeta}_g) = ||\boldsymbol{P}\mathbf{1}_N - \tilde{\boldsymbol{\zeta}}_{d}^t||_2^2 + ||\boldsymbol{P}'\mathbf{1}_I - \boldsymbol{\zeta}_{g}||_1.
\end{equation}
Instead of using KL divergence for both marginals, GL employs a mixed-norm approach for its discrepancy penalties. It uses the first term, a pixel-wise $L_2$ norm penalty on the source marginal, encouraging the reconstructed density to match the prediction. For the target marginal, it uses the second term, a point-wise $L_1$ norm penalty on the target marginal, to directly penalize the absolute error in the count at each annotation. The final GL loss is then the sum of this explicit discrepancy term and the transport cost (plus an entropy regularizer):
\begin{equation}
L_\text{GL} = \langle \mathbf{C}_\text{p}, \boldsymbol{P}_*^t \rangle - \epsilon H(\boldsymbol{P}_*^t) + D_\text{GL}(\tau(\tilde{\boldsymbol{\zeta}}_{d}^t),\boldsymbol{\zeta}_g),
\end{equation}
Here, GL~\cite{wan2021generalized} defines a perspective-guided cost matrix $\mathbf{C}_\text{p}$ to differ from DMCount~\cite{wang2020dm} and Ma et al.~\cite{ma2021learning}.
Meanwhile, unlike Ma et al.~\cite{ma2021learning}, which embeds discrepancy implicitly via KL penalties, GL defines its discrepancy terms explicitly and uses a mixed-norm ($L_2$ and $L_1$) strategy.

This overview illustrates the different strategies for defining and handling the transport discrepancy $D(\cdot)$, progressing from a strict global constraint to more flexible, localized penalties. Our GST framework, with its simple $L_1$ discrepancy loss, aligns most closely with the philosophy of penalizing the final transported mass difference, but achieves this with a pre-computed, stable transport kernel, avoiding the complexities of joint OT optimization.

\section{Proof of Theorem 1.}
\label{sec:supp_proof1}
\begin{thm}
For probability distributions $\boldsymbol{P}_X$ on $\mathcal{X}$ and $\boldsymbol{P}_Y$ on $\mathcal{Y}$, there exists a transport plan $\hat{\boldsymbol{P}} \in \mathcal{U}(\boldsymbol{P}_X, \boldsymbol{P}_Y)$ that can be expressed as $\hat{\boldsymbol{P}} = \text{diag}(\boldsymbol{P}_X) \cdot \boldsymbol{\mathcal{K}}$. Here, $\boldsymbol{\mathcal{K}}$, termed transport kernel, has elements defined as: 
\begin{equation}
\boldsymbol{\mathcal{K}}_{i,n}=\frac{P(x_i|y_n) P_Y(y_n)}{\sum_{n=1}^N P(x_i|y_n) P_Y(y_n)}.
\end{equation} 
\label{theorem:main} 
\end{thm}

\begin{framedproof}{Proof of Theorem 1}
Considering that a transport plan is fundamentally a joint probability distribution with the source and target distributions as marginals, we construct a transport plan $\hat{\boldsymbol{P}}$ by defining each element $\hat{P}_{i,n}$ as the joint probability $P(x_i, y_n)$. 
By the chain rule of probability, this joint probability can be decomposed as $P(x_i, y_n) = P(y_n|x_i) P_X(x_i)$, where $P_X(x_i)$ is the marginal probability from the source distribution $\boldsymbol{P}_X$. This decomposition naturally leads to the matrix structure $\hat{\boldsymbol{P}} = \text{diag}(\boldsymbol{P}_X) \cdot \boldsymbol{\mathcal{K}}$, where the transport kernel $\boldsymbol{\mathcal{K}}$ has elements $\boldsymbol{\mathcal{K}}_{i,n} = P(y_n|x_i)$.

To obtain the computational form of the kernel, we apply Bayes' theorem to $P(y_n|x_i)$:
\begin{equation}
P(y_n|x_i) = \frac{P(x_i|y_n) P_Y(y_n)}{P_X(x_i)}.
\label{eq:bayes_supp}
\end{equation}
The denominator $P_X(x_i)$ can be expanded using the law of total probability by summing over all possible states $y_n \in \mathcal{Y}$:
\begin{equation}
P_X(x_i) = \sum_{n=1}^{N} P(x_i|y_n) P_Y(y_n).
\label{eq:total_prob_supp}
\end{equation}
Substituting Eq.~\ref{eq:total_prob_supp} into the denominator of Eq.~\ref{eq:bayes_supp} yields the exact form of the kernel element as stated in the theorem:
\begin{equation}
\boldsymbol{\mathcal{K}}_{i,n} = \frac{P(x_i|y_n) P_Y(y_n)}{\sum_{n=1}^{N} P(x_i|y_n) P_Y(y_n)}.
\label{eq:final_kernel_form_supp}
\end{equation}

Now, we verify that $\hat{\boldsymbol{P}}$, constructed with this kernel, is a valid transport plan in the admissible set $\mathcal{U}(\boldsymbol{P}_X, \boldsymbol{P}_Y)$. First, we substitute the derived kernel form back into the expression for $\hat{P}_{i,n}$:
\begin{equation}
\begin{aligned}
\hat{P}_{i,n} &= P_X(x_i) \cdot \boldsymbol{\mathcal{K}}_{i,n} \\
&= P_X(x_i) \cdot \frac{P(x_i|y_n) P_Y(y_n)}{P_X(x_i)} \\
&= P(x_i|y_n) P_Y(y_n) = P(x_i, y_n).
\end{aligned}
\label{eq:p_hat_is_joint_supp}
\end{equation}
This confirms that our constructed $\hat{P}_{i,n}$ is indeed the joint probability. With this established, verifying the marginal constraints becomes straightforward.

The source marginal constraint ($\hat{\boldsymbol{P}}\mathbf{1}_N = \boldsymbol{P}_X$) is satisfied because summing each row $i$ yields:
\begin{equation}
\sum_{n=1}^{N} \hat{P}_{i,n} = \sum_{n=1}^{N} P(x_i, y_n) = P_X(x_i),
\end{equation}
by the law of total probability.

Similarly, the target marginal constraint ($\hat{\boldsymbol{P}}'\mathbf{1}_I = \boldsymbol{P}_Y$) is satisfied because summing each column $n$ yields:
\begin{equation}
\sum_{i=1}^{I} \hat{P}_{i,n} = \sum_{i=1}^{I} P(x_i, y_n) = P_Y(y_n),
\end{equation}
also by the law of total probability.

Since $\hat{P}_{i,n} = P(x_i, y_n) \ge 0$, the non-negativity condition also holds. Therefore, the constructed matrix $\hat{\boldsymbol{P}}$ is a valid transport plan in $\mathcal{U}(\boldsymbol{P}_X, \boldsymbol{P}_Y)$. This completes the proof.
\end{framedproof}

According to Theorem 1 and its proof, we have constructed a transport plan that operates within a clear and interpretable probabilistic framework, representing the transport assignment as the simple product of a transport kernel and the source probability distribution.

\section{Implementation Details}
\label{sec:supp_implementation}
All experiments are conducted on a single NVIDIA RTX 4090 GPU using PyTorch 2.0.1 and CUDA 11.8. The 2D Gaussian Splatting (2DGS) process is implemented based on the gsplat library~\cite{ye2024gsplatopensourcelibrarygaussian}.  

\subsection{2DGS Pre-computation}
For the 2DGS optimization in all tasks, we use the Adam optimizer with a learning rate of 0.01. The deformity elimination hyperparameter $\delta$
(from Eq. 10 in the main manuscript) is set to 1.5, and the shape loss weight $\beta$ is set to 0.2. The number of optimization iterations for 2DGS is 4,000 and 1,000 for counting datasets and landmark datasets, respectively. The one-time 2DGS pre-computation is highly efficient, typically taking 3-7 seconds per image on our setup, and remaining under 10 seconds even for extreme 4K images with over 4,500 points. For a more detailed breakdown of the costs across different scenes, we provide several representative examples in Table~\ref{tab:examples for 2DGS}.

\begin{table*}[ht]
    \centering
    \begin{tabular}{cc|cc|cc}
    \toprule[1.0pt]
         \multirow{2}{*}{Resolution} & \multirow{2}{*}{Num. of Annotations} & \multicolumn{2}{c|}{Rasterization Time~(s)} & \multicolumn{2}{c}{Backward Time~(s)} \\
         & & Total & Per Step & Total & Per Step \\
         \hline
         1920 $\times$ 960  & 953  & 0.690 & 0.00017 & 1.784 & 0.00045 \\
         1920 $\times$ 1264 & 489  & 0.900 & 0.00022 & 2.198 & 0.00055 \\
         1920 $\times$ 1264 & 8359 & 0.959 & 0.00024 & 3.482 & 0.00087 \\
         2500 $\times$ 1667 & 2980 & 1.159 & 0.00029 & 4.242 & 0.00106 \\
         2731 $\times$ 2048 & 1072 & 1.555 & 0.00039 & 5.380 & 0.00134 \\
         4110 $\times$ 1744 & 4535 & 1.316 & 0.00033 & 7.969 & 0.00199 \\
    \bottomrule[1.0pt]
    \end{tabular}
    \caption{Detailed breakdown of the 2DGS pre-computation time on an NVIDIA 4090 GPU. The costs are reported for images with varying resolutions and numbers of annotations. All examples are run for 4,000 optimization iterations.}
    \label{tab:examples for 2DGS}
\end{table*}

\subsection{Crowd Counting} 
\begin{figure*}[h!]
\centering
\includegraphics[width=\textwidth]{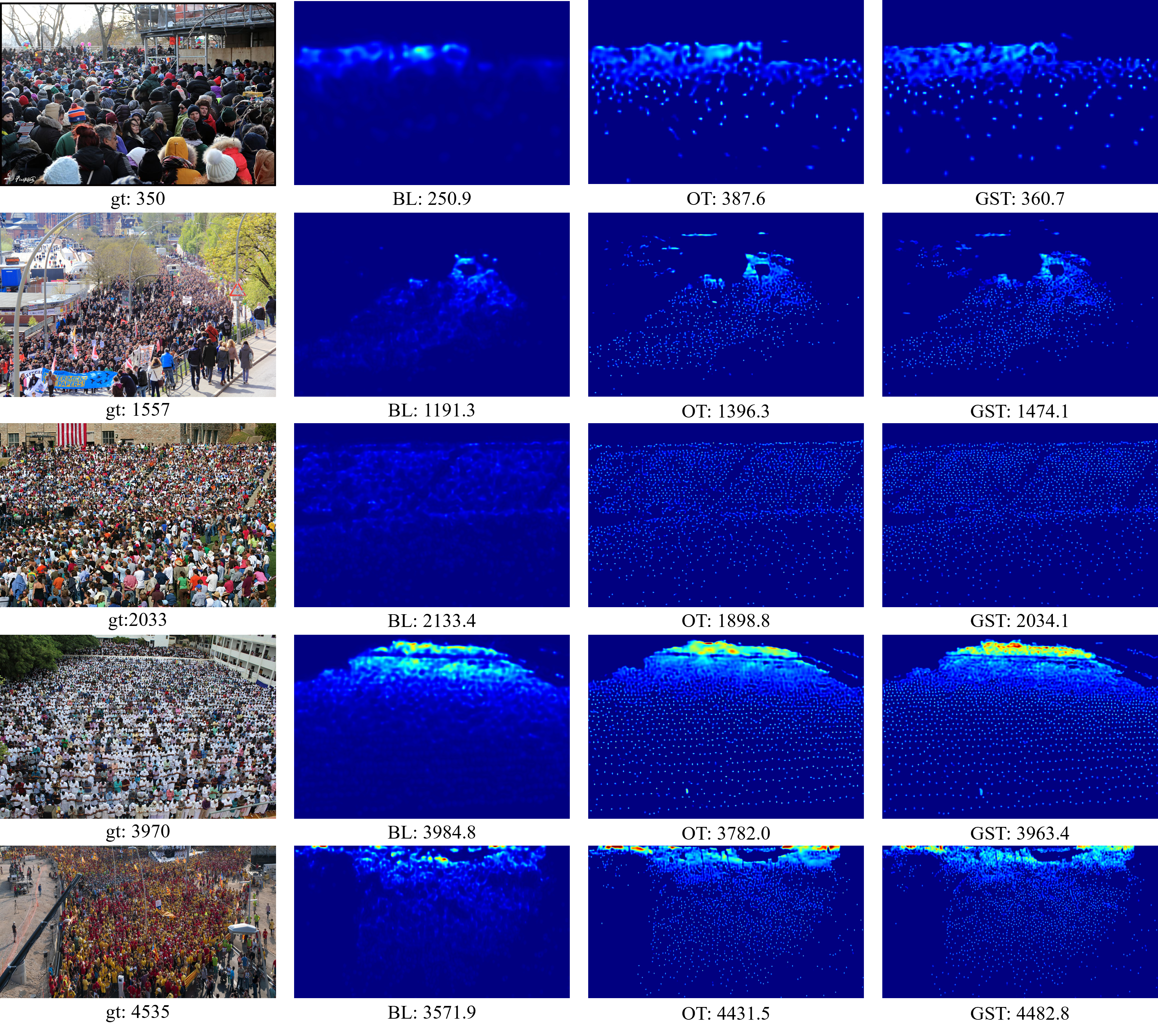}
\caption{Additional visualization results for crowd counting. From left to right are the input image, the estimated density maps of BL, OT, and our GST. }
\label{fig:supp_vis_crowd}
\end{figure*}

To validate the versatility of our method, we use two backbones~(as in Table 5 of \emph{Manuscript}): a VGG-19~\cite{simonyan2014vgg} with ImageNet pre-trained weights, and a Transformer architecture following Lin et al.~\cite{lin2024gramformer}. The cut-off threshold $d$ for pixel-to-background correspondence is set to 3. 
For model training, we use the Adam optimizer with a batch size of 1. The learning rate is set to $1e-5$ for VGG-19 and $5e-6$ for the Transformer. The image crop sizes are $ 384 \times 384$ for NWPU and $512 \times 512$ for other benchmarks. Additional visualization results are provided in Fig.~\ref{fig:supp_vis_crowd}.

\subsection{Landmark Localization} We evaluate the performance of landmark localization on MPII~\cite{andriluka14cvpr}. The dataset contains 2.5k annotated images with over 40k human instance. Each subject is annotated with 16 anatomical keypoints, including challenging cases featuring severe occlusions (23\% of instances), extreme kinematic configurations, and multi-person interactions. 
Following the standard evaluation protocol~\cite{andriluka14cvpr}, we report the head-normalized Probability of Correct Keypoint (PCKh@0.5) metric, where a detection is considered correct if the predicted joint location lies within 50\% of the head segment length from the ground truth. HRNet~\cite{sun2019hrnet} is established as our baseline framework, evaluating both HRNet-W32 and HRNet-W48 backbones.  All input preprocessing, density map postprocessing, and training protocols align with the baseline. For Gaussian rendering, we first preprocess the inputs as in~\cite{sun2019hrnet} to extract subject regions. The optimization runs 1,000 iterations with a learning rate of 0.01. For transport plan construction, we formulate dedicated foreground and background transport plans between each joint and its corresponding density map channel. The cut-off distance is 3. The learning rate of model training is $10^{-4}$.

\section{In-Depth Comparison Between OT and GST}
\label{sec:supp_comparison}

The proposed GST framework fundamentally differs from OT-based methods~\cite{wang2020dm, ma2021learning} in two key aspects: \textbf{the optimization paradigm} and \textbf{the nature of the spatial transport representation}.

\subsection{Optimization Paradigm}

The optimization in OT-based methods follows a \textbf{bi-level} structure. At each training iteration $t$, the process is:
\begin{align}
    \boldsymbol{P}_*^t &= \arg\min_{\boldsymbol{P} \in \mathcal{U}(\tilde{\boldsymbol{P}_X^t}, \boldsymbol{P}_Y)} \langle \mathbf{C}, \boldsymbol{P} \rangle  \quad &&\text{\# Inner loop} \label{eq:supp_inner_loop} \\
    \theta_{t+1} &= \text{optimizer}(\theta_t, \nabla_{\theta} \langle \mathbf{C}, \boldsymbol{P}_*^t \rangle) \quad &&\text{\# Outer loop} \label{eq:supp_outer_loop}
\end{align}
where $\tilde{\boldsymbol{P}_X^t} = f(\boldsymbol{\zeta}_{img};\theta_t) / ||f(\boldsymbol{\zeta}_{img};\theta_t)||_1$ is the normalized density predicted by the network with parameters $\theta_t$, $\boldsymbol{\zeta}_{img}$ is the input image, $\boldsymbol{P}_Y=\boldsymbol{\zeta}_{g} / ||\boldsymbol{\zeta}_{g}||_1$ is its normalized annotation map, and $\mathbf{C}$ is the pair-wise cost matrix. Note that for clarity in demonstrating the bi-level structure, we have presented the objective function using only the transport cost term 
$W(\boldsymbol{P}_*^t)=\langle \boldsymbol{C},\boldsymbol{P}_*^t \rangle$. This corresponds to the general loss form in Eq.~2 of the \emph{Manuscript} where the transport discrepancy term is omitted for simplicity of notation.

Eq.~\ref{eq:supp_inner_loop} shows the inner loop that aims to solve for the optimal plan given the current estimated density map and ground truth annotations. And Eq.~\ref{eq:supp_outer_loop} is the outer loop to update network weights with $\text{optimizer}(\cdot,\cdot)$ like Adam. Therefore, the transport plan and network parameters are optimized through nested loops. This requires computationally expensive recomputation of the transport plan (typically via Sinkhorn~\cite{cuturi2013sinkhorn} iterations) at each parameter update, leading to $\mathcal{O}(kpq)$ complexity per training iteration for $k$ Sinkhorn iterations, $p$ annotated points and $q$ pixels.

In contrast, the proposed GST framework establishes a \textbf{single-level optimization} process by decoupling the transport computation from network training. GST operates through a single optimization loop where only the network parameters $\theta$ are iteratively updated, while keeping the pre-computed transport kernel $\boldsymbol{\mathcal{K}}$ fixed. The training objective simplifies to:
\begin{equation}
    \theta^* = \arg\min_{\theta} ||\boldsymbol{\mathcal{K}}'f(\boldsymbol{\zeta}_{img};\theta) - \boldsymbol{\zeta}_{g}||_1
\end{equation}
where the transport kernel $\boldsymbol{\mathcal{K}}$ is pre-computed based on the pixel-to-annotation correspondence, as introduced in Section~3 of the \emph{Manuscript}, before training begins. By fixing the transport kernel, GST eliminates the computational overhead of repeated OT solves, reducing the optimization to a single-level process and the theoretical time complexity of the transport computation during training to $\mathcal{O}(1)$. The empirical results in Table~7 of the \emph{Manuscript} demonstrate that GST preserves accuracy while significantly improving training efficiency.

\subsection{Spatial Characteristic of Transport} 
From the perspective of probability measure transportation, the transport kernel $\boldsymbol{\mathcal{K}}$ intrinsically encodes the \textbf{spatial relationship} between the target distribution $\boldsymbol{\zeta}_g$ and the source distribution within the transport plan $\hat{\boldsymbol{P}}=\text{diag}(\boldsymbol{P}_X)\boldsymbol{\mathcal{K}}$. The elements $\boldsymbol{\mathcal{K}}_{i,n}$ quantify the ratio of source mass at location $x_i \in \mathcal{X}$ transported to the target $y_n \in \mathcal{Y}$, formally establishing a discrete spatial mapping from the image coordinate space $\mathcal{X}$ to the annotation space $\mathcal{Y}$. This matrix embodies two fundamental spatial representations:
\begin{itemize}[noitemsep,topsep=0pt,parsep=2pt,partopsep=0pt,leftmargin=1em]
    \item Each column vector $\boldsymbol{\mathcal{K}}_{:,n} \in \mathbb{R}^I$ acts as a spatial basis function, defining the spatial influence domain of the $n$-th annotation point over the entire image. 
    \item Each row vector $\boldsymbol{\mathcal{K}}_{i,:} \in \mathbb{R}^N$ represents the affinity distribution of the $i$-th pixel across all target annotations.
\end{itemize}

For OT-based methods, an analogous transport kernel can be conceptually defined as $\boldsymbol{\mathcal{K}}^t = \text{diag}(\tilde{\boldsymbol{P}_X^t})^{-1} \boldsymbol{P}_*^t$. This OT-derived kernel is inherently \textbf{dynamic}, as it depends on the network's prediction $\boldsymbol{P}_X^t$ at each iteration $t$. As the network learns, both the predicted density map and this implicit kernel change, which can lead to instability in the transport's spatial structure. 

In contrast, our GST framework employs pre-computed, spatially grounded kernel matrices that maintain a \textbf{stable spatial relationship} throughout training. As formalized in Eq.~4 of the \emph{Manuscript}, the GST kernel $\boldsymbol{\mathcal{K}}$ depends solely on the conditional likelihood $P(x|y)$, which is derived from the input image's appearance characteristics via Gaussian Splatting. Importantly, $\boldsymbol{\mathcal{K}}$ remains invariant to the predicted density values, being fundamentally determined by the underlying spatial relationships. This formulation offers two key advantages. First, it stabilizes the supervision signal, allowing the task-specific network to focus solely on refining density predictions against a fixed target mapping. Second, through the parameterization of the underlying Gaussians (as in Eqs.~8, 10, and 12 of the \emph{Manuscript}), our kernel explicitly encodes and preserves local geometric properties such as shape and orientation. In contrast to the fluctuating target of OT, GST provides a consistent and geometrically-informed spatial supervision throughout the training process.

\clearpage
\bibliography{aaai2026}